\documentclass[10pt,twocolumn,letterpaper]{article}

\usepackage[pagenumbers]{cvpr} 

\definecolor{cvprblue}{rgb}{0.21,0.49,0.74}
\usepackage[pagebackref,breaklinks,colorlinks,allcolors=cvprblue]{hyperref}

\def\confName{CVPR}
\def\confYear{2027}

\title{Motion Conditioned World Modeling for Embodied Visual Tracking}

\author{
    Jiahao Zhang\textsuperscript{1} \quad
    Xuan Yao\textsuperscript{1,2} \quad
    Junyu Gao\textsuperscript{1,2}\thanks{Corresponding author.} \quad
    Changsheng Xu\textsuperscript{1,2,3} \\
    \textsuperscript{1}State Key Laboratory of Multimodal Artificial Intelligence Systems (MAIS), \\ Institute of Automation, Chinese Academy of Sciences, Beijing, China\\
    \textsuperscript{2}School of Artificial Intelligence, University of Chinese Academy of Sciences (UCAS), Beijing, China\\
    \textsuperscript{3}Peng Cheng Laboratory, Shenzhen, China\\
    {\tt\small jahaozhang@163.com, yaoxuan2022@ia.ac.cn, junyu.gao@nlpr.ia.ac.cn, csxu@nlpr.ia.ac.cn} 
}

\usepackage{amsmath}
\usepackage{amssymb}
\usepackage{mathtools}
\usepackage{amsthm}

\usepackage{textcomp}
\usepackage{caption}
\usepackage[table,xcdraw]{xcolor}
\usepackage{multirow,booktabs,makecell}
\usepackage{setspace}
\usepackage{threeparttable}
\usepackage{float}
\usepackage{stfloats}
\usepackage{gensymb}
\usepackage{comment}
\usepackage{algorithm}
\usepackage{algorithmicx}
\usepackage{algpseudocode}
\usepackage{arydshln}
\usepackage{url}
\usepackage{soul}   
\soulregister\cite7
\soulregister\ref7
\usepackage{tabularx}
\definecolor{Gray}{gray}{0.5}   
\usepackage{xcolor,colortbl}
\usepackage{adjustbox}
\usepackage{enumitem}
\usepackage{pifont}
\usepackage{upquote}  
\usepackage{hhline}
\usepackage{threeparttable}
\usepackage[table]{xcolor}
\definecolor{myblue}{RGB}{0,100,200}
\definecolor{mygreen}{HTML}{009900}
\definecolor{mygray}{gray}{0.5}
\definecolor{mypurple}{HTML}{CDA0CD}

\definecolor{BestColor}{HTML}{ffffff}
\definecolor{SecondColor}{HTML}{ffffff}

\begin{document}
\maketitle
\begin{abstract}
World models support embodied decision-making by predicting future environment states.
In embodied visual tracking, uncertain target motion and robot-induced viewpoint changes jointly affect visual observations, making future-state prediction challenging.
We propose Motion-Conditioned World Modeling, which incorporates predicted target-motion distributions and robot ego-motion into latent state transitions.
A Geometry--Progress module jointly models geometric paths and single-step travel distances through shared motion modes and path-conditioned progress distributions.
This formulation captures uncertainty in route selection and motion progress while preserving their dependence.
Action-induced coordinate transformations express predicted paths in the next robot frame, providing geometrically grounded conditions for future visual-state prediction.
Future visual feature reconstruction supervises the motion predictor through the conditioned transition branch, linking explicit motion representations to subsequent observations.
The model is jointly optimized with motion likelihood, feature reconstruction, and expert-action supervision, enabling the tracking policy to combine visual context with predictive motion cues.
Experiments on three embodied tracking benchmarks demonstrate improved tracking performance and cross-environment generalization with a lightweight $0.5$B backbone.
\end{abstract}    
\section{Introduction}
\label{sec:introduction}
\begin{figure}[t]
    \centering
    \includegraphics[width=\linewidth]{
        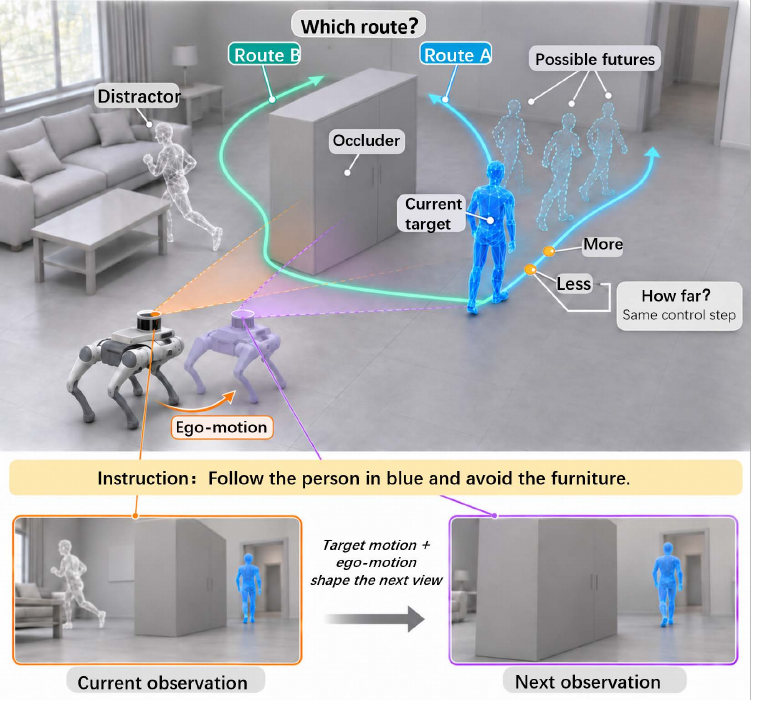
    }
    \caption{
Uncertain geometric routes and single-step progress yield different target-motion hypotheses, while robot ego-motion changes the observation frame.
Their joint effects shape future observations, motivating motion-conditioned world-state prediction for tracking.
Dashed figures represent alternative futures of the same target.
}
    \label{fig:introduction}
\end{figure}
Embodied Visual Tracking (EVT) is an important task in embodied intelligence that requires a mobile robot to continuously perceive, identify, and follow a designated target in a dynamic environment while actively adjusting its viewpoint through its own motion to maintain stable tracking.
This task tightly couples target perception with robot motion control and provides a fundamental capability for service robotics, companion robotics, and human--robot collaboration.

Unlike conventional visual object tracking, which primarily relies on passive visual observations, EVT requires the robot to understand the dynamics of both the target and its environment while actively controlling its motion to keep the target observable.
The task therefore involves a tight coupling among target motion modeling, dynamic environment perception, and robot decision-making.
However, target turns, occlusions, and interactions with distractors can lead to missing observations and perceptual ambiguity.
Meanwhile, the robot's own translation and rotation alter the target's position and appearance in its field of view.
Reliable tracking therefore requires estimating how the target moves and anticipating how robot actions affect subsequent observations.
World models integrate historical observations into latent states and learn action-conditioned state evolution, offering a suitable framework for addressing these challenges.
By linking current perception to future prediction, they provide predictive information for tracking through occlusions and viewpoint changes.

However, existing world models primarily construct latent states from visual observations and rely on learned state transitions to predict future environmental changes.
When applied to EVT, this reliance on implicit visual representations for dynamics modeling has two limitations.
First, target motion and robot ego-motion have distinct physical meanings: the target's movement, turning, and motion progress determine its future spatial position, whereas changes in the robot's pose alter the viewpoint and the spatial relationship between the robot and the target.
Both forms of motion jointly shape consecutive visual observations, and learning state transitions solely from visual features makes it difficult to fully exploit the physically meaningful information underlying these dynamics.
Second, future target motion is uncertain, particularly when the target turns, becomes occluded, or is affected by environmental disturbances.
A single predicted future position cannot fully capture the different motion behaviors the target may exhibit.
These limitations restrict the ability of world models to predict complex dynamics in embodied tracking scenarios and support subsequent tracking decisions.

To address these limitations, we propose \emph{Motion-Conditioned World Modeling} for EVT, which uses motion information with explicit geometric meaning to guide the evolution of latent world states.
To account for the coupling between target motion and robot ego-motion, we explicitly represent the geometric relationship between target displacement and changes in the observation coordinate frame, enabling the world model to predict future states under their combined effects.
To capture uncertainty in future target motion, we model a joint distribution over the target's route choice and motion progress, incorporating multiple plausible motion outcomes into future prediction.
Furthermore, we link motion modeling to future observations, allowing the learned motion representations to be supervised by both motion annotations and subsequent observations.
The resulting motion information provides predictive guidance for the tracking policy.

Our main contributions are as follows:


\begin{itemize}


    \item We propose a motion-conditioned world model for EVT that transforms predicted target-path distributions into the next robot frame using action-induced ego-motion, providing geometric conditions for latent state transitions.

\item We introduce Geometry--Progress joint probabilistic modeling of geometric paths without temporal indexing and single-step travel distances. Shared mixture modes and path-conditioned progress distributions capture their dependence and uncertainty.

\item We conduct extensive experiments on EVT-Bench, Gym-UnrealCV, and UnrealZoo, demonstrating improved tracking performance and cross-environment generalization with a $0.5$B backbone. Ablation studies validate the key design choices, while zero-shot deployment on a Unitree GO2 robot further demonstrates sim-to-real transfer.
\end{itemize}

\section{Related Work}
\label{sec:related_work}

\noindent\textbf{World Models.}
World models learn environment dynamics for prediction and control.
Early studies learn action-conditioned transitions over compact latent states, supporting planning or policy optimization through imagined trajectories~\cite{ha2018worldmodels,hafner2019planet,hafner2025dreamerv3}.
Recent approaches predict future representations in pretrained visual feature spaces to capture control-relevant dynamics~\cite{zhou2025dinowm,assran2025vjepa2}.
Another line generates future images or videos to simulate action consequences and guide decision-making~\cite{bar2025nwm}.
In embodied tracking, future observations jointly reflect target motion and robot-induced viewpoint changes.
Tracking-oriented world models improve prediction using structured scene and target states or constraints on action-dependent information flow~\cite{yao2026bipartite,hu2026cstwm}.
Our approach explicitly couples target-motion uncertainty with changes in the observer's coordinate frame.
Using action-induced ego-motion, we transform predicted geometric paths into the next robot frame, providing geometrically defined motion conditions for latent state transitions.

\noindent\textbf{Embodied Visual Tracking.}
Embodied visual tracking couples target identification with autonomous control to maintain visibility and an appropriate following distance.
Early methods learn tracking policies through end-to-end reinforcement learning and multi-agent interaction~\cite{luo2018active,zhong2021distraction}.
Language-conditioned approaches connect instruction understanding with low-level control through hierarchical spatial goals~\cite{wu2025HIEVT}.
TrackVLA jointly learns target recognition and tracking trajectory generation using shared vision-language representations~\cite{wang2025trackvla}, while TrackVLA++ introduces spatial reasoning and long-term memory to improve tracking under occlusion and interference~\cite{liu2025trackvla++}.
CoMaTrack introduces competitive multi-agent reinforcement learning, training VLA trackers against adaptive opponents to improve planning and robustness to interference~\cite{liu2026comatrack}.
Occlusion-aware planning supports target recovery~\cite{sun2026oavat}, and embodiment-adaptive methods use interaction history to accommodate platform-specific control constraints~\cite{wu2026adatracker,guo2026phaselora}.
Our approach connects probabilistic motion prediction with world-state learning, allowing the tracking policy to combine current visual context with predictive motion cues.

\noindent\textbf{Probabilistic Motion Prediction.}
Future target motion is inherently multimodal, as the same history may lead to different route choices and amounts of progress.
Conditional generative methods model multiple possible futures, incorporating agent interactions, scene information, or kinematic models to balance prediction diversity and motion plausibility~\cite{gupta2018socialgan,salzmann2020trajectronpp}.
Spatially structured methods use endpoints, waypoints, and paths as hierarchical conditions for goal- or route-guided prediction~\cite{mangalam2020pecnet,mangalam2021ynet}.
Incorporating road topology further separates route selection from motion along a route, decomposing different sources of uncertainty~\cite{deo2022pgp}.
Building on these ideas, we learn a joint distribution over geometric paths without temporal indices and arc-length progress within a single control interval.
This representation preserves the dependence between path geometry and motion progress and, together with robot ego-motion, conditions world-state transitions.
Future visual feature reconstruction provides additional supervision from subsequent observations, connecting probabilistic motion prediction with world-model learning for embodied tracking.

\begin{figure*}[t]
    \centering
    \includegraphics[width=\textwidth]{
        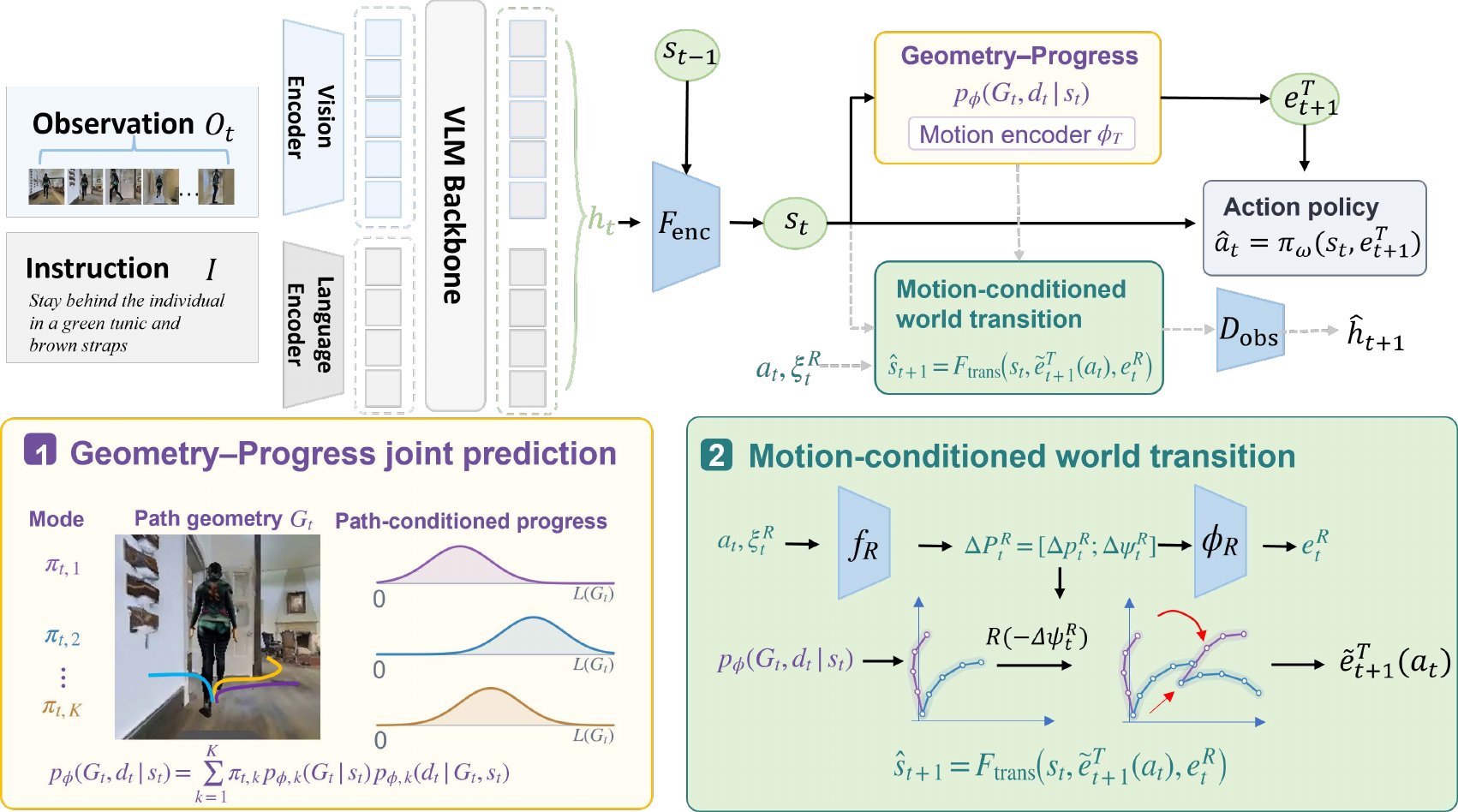
    }
    \caption{
\textbf{Overview of Motion-Conditioned World Modeling.}
A VLM and history encoder maintain a shared latent state from observations and instructions.
\textbf{(1)} Geometry--Progress jointly models geometric paths and single-step travel distance, capturing their dependence and uncertainty.
\textbf{(2)} The target-motion distribution is expressed in the next robot frame and combined with action-induced ego-motion to predict the next latent state.
Future visual feature reconstruction supervises both motion prediction and state transitions.
The policy uses the current state and predicted target motion to generate tracking actions.
Gray dashed arrows indicate training-only computations.
}
    \label{fig:framework}
\end{figure*}

\section{Method}
\label{sec:method}

We model embodied visual tracking (EVT) with a latent world model.
The robot must identify and follow a specified target from sequential visual observations while adjusting its viewpoint through its own motion.
Since the robot receives only local observations, we maintain a recurrent latent world state that summarizes visual history and supports target motion prediction and action generation.

Under the assumptions of planar robot motion and target motion along a geometric path, path geometry specifies a possible spatial route, while single-step travel distance determines progress along that route.
Together, they define a future target position in a fixed reference frame.
The translation and yaw change induced by a robot action then map this position into the next robot coordinate frame.
These quantities specify the kinematic relation between target motion and observer motion over one control interval.
We use them as explicit conditions for world-state transitions, while the latent visual state represents scene structure, target appearance, and occlusion.

Visual history contains motion cues, but the latent state does not explicitly specify separately supervised path geometry, motion progress, or their uncertainty.
Under occlusion or at a path junction, similar observations may admit different routes.
Even along the same route, different travel distances lead to different future positions.
Collapsing these possibilities into a single prediction may overlook plausible routes or produce an intermediate position, biasing the estimated target-relative position and consequently affecting steering and following-distance control.
We therefore jointly model path geometry and single-step progress to represent alternative routes and the corresponding progress distributions.
Motion labels assign explicit geometric meanings to these variables, and future visual feature reconstruction provides an additional training signal through the motion-conditioned transition branch.

Given the current observation $O_t$ and tracking instruction $I$, a pretrained vision-language model (VLM) extracts visual semantic features, and a history encoder updates the world state:
\begin{equation}
\begin{aligned}
h_t &= \psi_{\mathrm{VLM}}(O_t,I)
      \in \mathbb{R}^{d_h}, \\
s_t &= F_{\mathrm{enc}}(s_{t-1},h_t)
      \in \mathbb{R}^{d_s}.
\end{aligned}
\label{eq:world_state_encoding}
\end{equation}
Here, $h_t$ aggregates the VLM output tokens for the current observation, whereas $s_t$ incorporates historical information; $d_h$ and $d_s$ denote their feature dimensions.
From $s_t$, the model predicts a joint distribution over target paths and single-step progress.
This distribution is encoded into a motion representation and combined with action-induced robot ego-motion to predict the next world state.
The tracking policy also uses the predicted motion representation.
The training and inference procedures are described in Sec.~\ref{sec:overall_objective}.

\subsection{Motion-Conditioned World Modeling}
\label{sec:motion_conditioned_modeling}

We first define a probabilistic representation of target path geometry and motion progress, and then describe how it conditions world-state transitions together with robot ego-motion.

\noindent\textbf{Probabilistic Target Dynamics Modeling.}
We represent the target's geometric path $G_t$ in the fixed current robot frame $\mathcal{R}_t$.
The path starts at the current target position $g_{t,0}$ and continues through $J$ ordered spatial anchors describing its shape.
Each node $g_{t,j}$ is specified by a bearing angle $\theta_{t,j}$, measured from the robot's forward direction, and a radial distance $r_{t,j}$.
The node indices indicate spatial ordering along the path and do not specify future arrival times.

The travel distance $d_t$ describes the target's arc-length progress along the path during one control interval.
It measures single-step movement rather than the total path length or the straight-line displacement between endpoints.
This distinction allows the same geometric path to admit different progress values, while the same travel distance can correspond to different paths.

The path nodes and single-step travel distance together form the structured target-motion variables.
Conditioned only on the current latent state $s_t$, the model predicts their joint distribution to capture the dependence and uncertainty of path geometry and motion progress.
To represent uncertainty in route selection and progress, we use a joint mixture distribution with $K$ motion modes:
\begin{equation}
\begin{aligned}
&p_\phi(G_t,d_t\mid s_t) \\
&\quad=\sum_{k=1}^{K}\pi_{t,k}\,
p_{\phi,k}(G_t\mid s_t)\,
p_{\phi,k}(d_t\mid G_t,s_t),
\end{aligned}
\label{eq:joint_motion_distribution}
\end{equation}
where $\pi_{t,k}\geq 0$ and $\sum_{k=1}^{K}\pi_{t,k}=1$.
The shared mode and the path-conditioned progress distribution capture the dependence between route geometry and travel distance.
Let $L(G_t)$ denote the total path arc length, computed after converting the polar nodes to Cartesian coordinates.
The progress distribution has support $0\leq d_t\leq L(G_t)$, so the predicted path must cover the target's movement over the next control interval.

To pass the motion distribution to downstream modules, we encode the mode probabilities and distribution parameters:
\begin{equation}
e_{t+1}^{T}
=\phi_T\!\left(
\{[\pi_{t,k};\eta_{t,k}]\}_{k=1}^{K}
\right)\in\mathbb{R}^{d_e}.
\label{eq:target_motion_encoding}
\end{equation}
Here, $\eta_{t,k}\in\mathbb{R}^{d_\eta}$ contains the path and progress distribution parameters of the $k$-th joint component.
The encoder maps each component to a motion token and aggregates these tokens into a $d_e$-dimensional representation.
The structured variable $x_t^T$ has explicit geometric meaning, whereas $e_{t+1}^T$ is a neural encoding of its predictive distribution, including mode weights and uncertainty parameters.
The next-step embedding $e_{t+1}^T$ is computed at time $t$, before conditioning on the robot action.

\noindent\textbf{Motion-Conditioned World State Transition.}
Target motion changes the target position in a fixed reference frame, while robot ego-motion changes the observation frame.
We account for these effects before applying the latent state transition by combining the predicted target motion with the pose change induced by the robot action.

For an action $a_t\in\mathbb{R}^{d_a}$, a robot motion model predicts a planar pose increment and its encoding:
\begin{equation}
\begin{aligned}
\Delta P_t^R
&=f_R(a_t;\xi_t^R)
 =[\Delta p_t^R;\Delta\psi_t^R], \\
e_t^R&=\phi_R(\Delta P_t^R)\in\mathbb{R}^{d_r}.
\end{aligned}
\label{eq:robot_ego_motion}
\end{equation}
Here, $\Delta P_t^R\in\mathbb{R}^{3}$ consists of a translation $\Delta p_t^R\in\mathbb{R}^{2}$ expressed in $\mathcal{R}_t$ and a yaw increment $\Delta\psi_t^R$.
The variable $\xi_t^R$ contains the current robot state and parameters required by the motion model.
The pose increment is predicted from the action, without access to the actual future robot pose.

Let $R(\cdot)$ denote a two-dimensional rotation matrix, and let $\Gamma_{G_t}(d_t)\in\mathbb{R}^{2}$ be the Cartesian target position obtained by arc-length interpolation along $G_t$ after converting its polar nodes to Cartesian coordinates.
For each predicted path-progress pair, the future target position in the next robot frame is
\begin{equation}
\widetilde p_{t+1}^T(a_t)
=R(-\Delta\psi_t^R)
\left[\Gamma_{G_t}(d_t)-\Delta p_t^R\right].
\label{eq:target_frame_transformation}
\end{equation}
In implementation, we apply the same rigid transformation to the Cartesian coordinates of all predicted path nodes and convert the transformed nodes back to polar coordinates, obtaining $\widetilde G_t(a_t)$.
The along-path progress $d_t$ remains unchanged under this transformation.
We then encode the transformed joint distribution using the same motion encoder to obtain the action-conditioned target representation $\widetilde e_{t+1}^T(a_t)$.
The world-state transition is
\begin{equation}
\widehat s_{t+1}
=F_{\mathrm{trans}}\!\left(
s_t,\widetilde e_{t+1}^T(a_t),e_t^R
\right)
\in\mathbb{R}^{d_s}.
\label{eq:conditioned_world_transition}
\end{equation}
The transition thus takes the current latent state, the target motion distribution expressed in the next robot frame, and the robot ego-motion encoding as inputs.
Its predicted state is subsequently decoded to reconstruct future visual features.

\subsection{Learning Objectives}
\label{sec:world_dynamics_learning}

We jointly learn the target motion distribution, the conditioned world-state transition, and the tracking policy.
Path labels, future observations, and expert actions are used during training and are not required at test time.

\noindent\textbf{Target Dynamics Learning.}
We construct geometric path labels $G_t^\ast$ and single-step progress labels $d_t^\ast$ from the training sequences.
The path labels are generated using a shortest-path algorithm, with all spatial nodes expressed in the fixed current robot frame $\mathcal{R}_t$.
The progress label is computed by summing the Euclidean distances between consecutive dense target positions within the current control interval.
All positions are expressed in a common fixed coordinate frame, ensuring that the accumulated distance reflects target movement independently of robot ego-motion.
We supervise the joint motion distribution using the negative log-likelihood over valid annotations:
\begin{equation}
\mathcal{L}_{\mathrm{mot}}
=-\mathbb{E}_t\!\left[
\log p_\phi(G_t^\ast,d_t^\ast\mid s_t)
\right].
\label{eq:motion_likelihood_loss}
\end{equation}
This objective encourages the model to assign higher probability to the annotated path-progress pairs and learn their statistical dependence.

\noindent\textbf{Future World State Reconstruction}. To train the conditioned state transition against subsequent observations, we decode the predicted state into the visual feature space:
\begin{equation}
\widehat h_{t+1}
=D_{\mathrm{obs}}(\widehat s_{t+1})
\in\mathbb{R}^{d_h}.
\label{eq:future_feature_decoding}
\end{equation}
We align it with the semantic features of the observed next frame using
\begin{equation}
\mathcal{L}_{\mathrm{wm}}
=\mathbb{E}_t\!\left[
\left\|\widehat h_{t+1}-h_{t+1}\right\|_2^2
\right].
\label{eq:world_reconstruction_loss}
\end{equation}
For this reconstruction branch, $a_t$ is the recorded action paired with the observed transition, while the target motion conditions are obtained from the predicted distribution.
Ground-truth future paths and travel distances are not substituted for the predicted motion conditions.
Reconstruction gradients propagate through $F_{\mathrm{trans}}$ and the motion encoders to the differentiable distribution-parameter prediction branches and the shared state encoder.
The reconstruction target is a visual semantic feature vector, so the model learns future dynamics in feature space.

\noindent\textbf{Overall Training Objective.}
\label{sec:overall_objective}
The tracking policy predicts the control action from the current state and the target motion representation computed before the action-dependent coordinate transformation:
\begin{equation}
\widehat a_t
=\pi_\omega(s_t,e_{t+1}^T)
\in\mathbb{R}^{d_a}.
\label{eq:tracking_policy}
\end{equation}
The action loss $\mathcal{L}_{\mathrm{act}}$ is computed as the $\ell_1$ loss between predicted actions $\widehat a_t$ and ground-truth expert actions $a_t^{\mathrm{GT}}$.
The overall training objective is
\begin{equation}
\mathcal{L}_{\mathrm{total}}
=\mathcal{L}_{\mathrm{mot}}
+\mathcal{L}_{\mathrm{wm}}
+\mathcal{L}_{\mathrm{act}}.
\label{eq:overall_training_loss}
\end{equation}
The motion likelihood learns the joint statistics of path geometry and progress, feature reconstruction connects the motion conditions to future observations, and action supervision trains tracking control.


\definecolor{mygray}{RGB}{128,128,128}
\begin{table*}[t]
    \caption{
    \textbf{Performance on EVT-Bench.} 
    The evaluation metrics are defined as follows: \textbf{Success Rate (SR)}, the proportion of episodes that the agent ends correctly oriented within 1–3m of the target; \textbf{Tracking Rate (TR)}, the proportion of timesteps with successful target tracking; and \textbf{Collision Rate (CR)}, the proportion of episodes terminated due to collisions with the target. 
    $\dag$: Uses GroundingDINO as the detector. 
    $\ddag$: Uses SoM~\cite{yang2023set} + GPT-4o~\cite{hurst2024gpt} as the visual foundation model. \textbf{Bold} indicates improvement over baselines, and \underline{underline} denotes the best result. * indicates experimental results that we have reproduced in this work.}
    \centering
    \resizebox{\textwidth}{!}{
    \begin{threeparttable}
    \begin{tabular}{lccccccccc}
        \toprule
        \multirow{2}{*}{Methods}& \multicolumn{3}{c}{\textit{Single-Target Tracking (STT)}} & \multicolumn{3}{c}{\textit{Distracted Tracking (DT)}} & \multicolumn{3}{c}{\textit{Ambiguity Tracking (AT)}} \\
        \cmidrule(lr){2-4} \cmidrule(lr){5-7} \cmidrule(lr){8-10}
        & SR$\uparrow$ & TR$\uparrow$ & CR$\downarrow$ & SR$\uparrow$ & TR$\uparrow$ & CR$\downarrow$ & SR$\uparrow$ & TR$\uparrow$ & CR$\downarrow$ \\
        \midrule
        IBVS$\dag$~\cite{gupta2016novel} & 42.9 & 56.2 & \underline{3.75} & 10.6 & 28.4 & \underline{6.14} & 15.2 & 39.5 & \underline{4.90} \\
        PoliFormer$\dag$~\cite{zeng2024poliformer} & 4.67 & 15.5 & 40.1 & 2.62 & 13.2 & 44.5 & 3.04 & 15.4 & 41.5 \\
        EVT~\cite{zhong2024EVT} & 24.4 & 39.1 & 42.5 & 3.23 & 11.2 & 47.9 & 17.4 & 21.1 & 45.6 \\
        EVT$\ddag$~\cite{zhong2024EVT} & 32.5 & 49.9 & 40.5 & 15.7 & 35.7 & 53.3 & 18.3 & 21.0 & 44.9 \\
        Uni-NaVid(Vicuna-7B)~\cite{zhang2024uni} & 25.7 & 39.5 & 41.9 & 11.3 & 27.4 & 43.5 & 8.26 & 28.6 & 43.7 \\
        {OpenTrackVLA(Qwen-0.6B)~\cite{opentrackvla2025}} & 64.8 & 84.4 & 5.00 & 33.6 & 66.3 & 8.84 & 39.6 & 76.7 & 6.38 \\
        OpenTrackVLA(Qwen-0.6B)*~\cite{opentrackvla2025} & 47.2 & 58.7 & 5.72 & 22.5 & 42.8 & 8.93 & 28.6 & 51.8 & 6.37 \\
        {BWM(Qwen2.5-0.5B)~\citep{yao2026bipartite}} & {86.3} & {84.8} & {3.90}
        &{36.5} & {56.6} & {10.8} & {52.6} & {69.7} & {6.10} \\
        \color{mygray}{USS (\cite{uss})} & \textcolor{mygray}{70.8} & \textcolor{mygray}{72.1} & \textcolor{mygray}{3.13} & \textcolor{mygray}{49.8} & \textcolor{mygray}{54.6} & \textcolor{mygray}{9.89} & \textcolor{mygray}{34.2} & \textcolor{mygray}{35.8} & \textcolor{mygray}{28.2} \\
\textcolor{mygray}{Qwen-RobotNav (Qwen3-VL-4B)~\citep{qwenrobotnav}} & \textcolor{mygray}{77.4} & \textcolor{mygray}{90.0} & \textcolor{mygray}{6.40} & \textcolor{mygray}{-} & \textcolor{mygray}{-} & \textcolor{mygray}{-} & \textcolor{mygray}{-} & \textcolor{mygray}{-} & \textcolor{mygray}{-} \\
        \textcolor{mygray}{TrackVLA(Vicuna-7B)~\cite{wang2025trackvla}} & \textcolor{mygray}{85.1} & \textcolor{mygray}{78.6} & \textcolor{mygray}{1.65} & \textcolor{mygray}{57.6} & \textcolor{mygray}{63.2} & \textcolor{mygray}{5.80} & \textcolor{mygray}{50.2} & \textcolor{mygray}{63.7} & \textcolor{mygray}{17.1} \\
        \textcolor{mygray}{NavFoM(Qwen2-7B)~\cite{zhang2025NavFoM}} & \textcolor{mygray}{85.0} & \textcolor{mygray}{80.5} & \textcolor{mygray}{-} & \textcolor{mygray}{61.4} & \textcolor{mygray}{68.2} & \textcolor{mygray}{-} & \textcolor{mygray}{-} & \textcolor{mygray}{-} & \textcolor{mygray}{-} \\
        \textcolor{mygray}{TrackVLA++(Qwen-7B)~\cite{liu2025trackvla++}} & \textcolor{mygray}{86.0} & \textcolor{mygray}{81.0} & \textcolor{mygray}{2.10} & \textcolor{mygray}{66.5} & \textcolor{mygray}{68.8} & \textcolor{mygray}{4.71} & \textcolor{mygray}{51.2} & \textcolor{mygray}{63.4} & \textcolor{mygray}{15.9} \\
        \textcolor{mygray}{VLingNav(LLaVA-Video-7B) (\cite{vlingnav})} & \textcolor{mygray}{88.4} & \textcolor{mygray}{81.2} & \textcolor{mygray}{2.07} & \textcolor{mygray}{67.6} & \textcolor{mygray}{73.5} & \textcolor{mygray}{5.51} & \textcolor{mygray}{-} & \textcolor{mygray}{-} & \textcolor{mygray}{-} \\
        
        \midrule
         \textbf{Qwen2.5-VLA-Adapter-0.5B~\cite{wang2025vlaadapter}} & 52.2 & 59.1 & 30.1 & 22.6 & 47.2 & 26.8 & 33.3 & 59.5 & 23.0 \\
        \rowcolor{mypurple!20}
        \textbf{Ours(Qwen2.5-0.5B)} & \underline{\textbf{87.9}} & \underline{\textbf{86.6}} & {\textbf{4.57}}
        &\underline{\textbf{41.2}} & \underline{\textbf{59.4}} & {\textbf{17.4}} & \underline{\textbf{55.4}} & \underline{\textbf{71.5}} & {\textbf{11.6}}  \\
        \bottomrule
    \end{tabular}          
	\end{threeparttable}
    \vspace{-1em}
    }
    \label{tab:evt-bench}
\end{table*}

\begin{table}[t]
    \caption{
    \textbf{Zero-shot Performance on Gym-UnrealCV.} 
    The evaluation metrics are defined as follows: \textbf{Episode Length (EL)}, the average number of steps before episode termination (maximum is 500); and \textbf{Success Rate (SR)}, the proportion of episodes completed for the full 500-step duration. 
    \textbf{Bold} indicates improvement over baselines, and \underline{underline} denotes the best result. * denotes results evaluated using official code.} 
    \centering
    \resizebox{\columnwidth}{!}{
    \begin{tabular}{lcccccc}
        \toprule
        \multirow{2}{*}{Methods} & \multicolumn{2}{c}{\textit{Single Target}} & \multicolumn{2}{c}{\textit{Distractor}} & \multicolumn{2}{c}{\textit{Unseen Objects}} \\
        \cmidrule(lr){2-3} \cmidrule(lr){4-5} \cmidrule(lr){6-7}
        & EL$\uparrow$ & SR$\uparrow$ & EL$\uparrow$ & SR$\uparrow$ & EL$\uparrow$ & SR$\uparrow$ \\
        \midrule
        DiMP~\cite{DiMP} & 367 & 0.58 & 309 & 0.27 & - & - \\
        SARL~\cite{SARL} & 394 & 0.57 & 240 & 0.14 & - & - \\
        AD-VAT~\cite{zhong2019ad} & 416 & 0.62 & 220 & 0.12 & - & - \\
        AD-VAT+~\cite{zhong2019ad+} & 454 & 0.76 & 224 & 0.12 & - & - \\
        TS~\cite{zhong2021TS} & 474 & 0.86 & 371 & 0.48 & - & - \\
        EVT~\cite{zhong2024EVT} & 490 & 0.95 & {459} & {0.81} & {480} & {0.96} \\
        OpenTrackVLA*~\cite{opentrackvla2025} & 465 & 0.82 & 436 & 0.76 & 232 & 0.35 \\
        BWM~\citep{yao2026bipartite} & \underline{500} & \underline{1.00} & 461 & 0.83 & \underline{500} & \underline{1.00} \\
        \textcolor{mygray}{TrackVLA~\cite{wang2025trackvla}} & \textcolor{mygray}{500} & \textcolor{mygray}{1.00} & \textcolor{mygray}{474} & \textcolor{mygray}{0.91} & \textcolor{mygray}{500} & \textcolor{mygray}{1.00} \\
        \textcolor{mygray}{TrackVLA++~\cite{liu2025trackvla++}} & \textcolor{mygray}{500} & \textcolor{mygray}{1.00} & \textcolor{mygray}{484} & \textcolor{mygray}{0.92} & \textcolor{mygray}{500} & \textcolor{mygray}{1.00}\\
        
        \midrule
        Qwen2.5-VLA-Adapter-0.5B~\citep{wang2025vlaadapter} & {494} & {0.98} & 397 & 0.64 & 460 & 0.71 \\
        \rowcolor{mypurple!20}
        \textbf{Ours} & \textbf{\underline{500}} & \textbf{\underline{1.00}} & \textbf{\underline{466}} & \textbf{\underline{0.84}} & \textbf{\underline{500}} & \textbf{\underline{1.00}} \\
        \bottomrule
    \end{tabular}
    }
    \label{tab:gym-unrealcv}
\end{table}

\begin{table*}[t]
    \caption{
    \textbf{Performance on UnrealZoo.} 
    Performance comparison across ten environments with dynamic instruction changes. Each cell reports \textbf{Average Accumulated Reward (AR), Average Episode Length (EL), and Success Rate (SR)} in the format \textbf{AR\,/\,EL\,/\,SR}. During each episode, the user instruction randomly changes four times, testing adaptation to shifting goals. \textbf{Bold} indicates improvement over baselines, and \underline{underline} denotes the best result. * denotes results evaluated using official checkpoints.}
    \centering
    \resizebox{\textwidth}{!}{
    \begin{tabular}{lcccccc}
        \toprule
        \multirow{2}{*}{Method} & \multirow{2}{*}{FPS} & \multicolumn{5}{c}{Environment Name} \\
        \cmidrule(lr){3-7}
        & & FlexibleRoom & Suburb & Supermarket & Parking Lot & Old Factory \\
        \midrule
        Bbox-based PID~\cite{wu2025HIEVT} & 50 & 154\,/\,365\,/\,0.50 & 124\,/\,386\,/\,0.36 & 175\,/\,298\,/\,0.26 & 112\,/\,356\,/\,0.38 & 80\,/\,309\,/\,0.30 \\
        Ensembled RL~\cite{zhong2024EVT} & 50 & 183\,/\,330\,/\,0.36 & 126\,/\,294\,/\,0.22 & 114\,/\,393\,/\,0.38 & 169\,/\,301\,/\,0.42 & 64\,/\,334\,/\,0.36 \\
        OpenVLA~\cite{kim2024openvla} & 8 & 123\,/\,253\,/\,0.25 & 63\,/\,311\,/\,0.18 & 68\,/\,233\,/\,0.10 & 12\,/\,205\,/\,0.08 & 5\,/\,226\,/\,0.12 \\
        GPT-4o~\cite{hurst2024gpt} & 0.38 & 15\,/\,194\,/\,0.10 & 14\,/\,241\,/\,0.20 & 54\,/\,286\,/\,0.16 & 23\,/\,286\,/\,0.30 & 5\,/\,297\,/\,0.32 \\
        VLM-Assistant (GPT-4o)~\cite{hurst2024gpt} & - & - & - & \underline{325}\,/\,\underline{493}\,/\,\underline{0.94} & 313\,/\,470\,/\,0.88 & \underline{315}\,/\,\underline{484}\,/\,\underline{0.92} \\
        HIEVT~\cite{wu2025HIEVT} & 50 & 278\,/\,500\,/\,1.00 & 166\,/\,445\,/\,0.72 & 212\,/\,422\,/\,0.64 & 169\,/\,491\,/\,0.93 & 128\,/\,469\,/\,0.84 \\
        OpenTrackVLA*~\cite{opentrackvla2025} & 5.6 & 285\,/\,475\,/\,0.89 & 202\,/\,341\,/\,0.33 & 293\,/\,430\,/\,0.64 & 313\,/\,435\,/\,0.69 & 288\,/\,422\,/\,0.61 \\
        BWM~\citep{yao2026bipartite} & 10 & {355}\,/\,\underline{500}\,/\,\underline{1.00} & {361}\,/\,{475}\,/\,{0.88} & {276}\,/\,{470}\,/\,{0.87} & \underline{342}\,/\,\underline{483}\,/\,\underline{0.94} & 297\,/\,473\,/\,0.91\\
        \midrule
        Qwen2.5-VLA-Adapter-0.5B~\cite{wang2025vlaadapter} & 10 & {251\,/\,\underline{500}\,/\,\underline{1.00}} & {249\,/\,426\,/\,0.64} & {187\,/\,406\,/\,0.63} & {220\,/\,483\,/\,0.93} & {212\,/\,462\,/\,0.80} \\
        \rowcolor{mypurple!20}
        \textbf{Ours} & 10 & \textbf{\underline{364}\,/\,\underline{500}\,/\,\underline{1.00}} & \textbf{\underline{363}\,/\,\underline{477}\,/\,\underline{0.89}} & {\textbf{265}\,/\,\textbf{473}\,/\,\textbf{0.89}} & {\textbf{{336}}\,/\,{480}\,/\,{0.93}} & \textbf{301\,/\,474\,/\,0.91}\\
        \midrule
        \multirow{2}{*}{Method} & \multirow{2}{*}{FPS} & \multicolumn{5}{c}{Environment Name} \\
        \cmidrule(lr){3-7}
        & & Container Yard & Desert Ruins & Brass Garden & Modular Old Town & Roof City \\
        \midrule
        Bbox-based PID~\cite{wu2025HIEVT} & 50 & 140\,/\,369\,/\,0.42 & 128\,/\,297\,/\,0.26 & 120\,/\,302\,/\,0.38 & 73\,/\,305\,/\,0.32 & 64\,/\,289\,/\,0.34 \\
        Ensembled RL~\cite{zhong2024EVT} & 50 & 27\,/\,327\,/\,0.24 & 56\,/\,368\,/\,0.34 & -8\,/\,348\,/\,0.34 & 14\,/\,356\,/\,0.32 & 19\,/\,322\,/\,0.38 \\
        OpenVLA~\cite{kim2024openvla} & 8 & 43\,/\,232\,/\,0.08 & 22\,/\,219\,/\,0.12 & 30\,/\,236\,/\,0.08 & -9\,/\,213\,/\,0.16 & 96\,/\,303\,/\,0.22 \\
        GPT-4o~\cite{hurst2024gpt}  & 0.38 & -43\,/\,149\,/\,0.00 & -5\,/\,249\,/\,0.26 & 3\,/\,243\,/\,0.18 & 29\,/\,304\,/\,0.28 & 18\,/\,256\,/\,0.20 \\
        VLM-Assistant (GPT-4o)~\cite{hurst2024gpt} & - & - & - & - & - & - \\
        HIEVT~\cite{wu2025HIEVT} & 50 & 156\,/\,469\,/\,0.76 & 148\,/\,434\,/\,0.74 & 126\,/\,425\,/\,0.72 & 85\,/\,433\,/\,0.64 & 86\,/\,400\,/\,0.58 \\
        OpenTrackVLA*~\cite{opentrackvla2025} & 5.6 & 297\,/\,432\,/\,0.70 & 296\,/\,427\,/\,0.74 & 206\,/\,390\,/\,0.43 & 290\,/\,436\,/\,0.72 & \underline{283}\,/\,414\,/\,0.66 \\
        BWM~\citep{yao2026bipartite} & 10 & {366}\,/\,\underline{500}\,/\,\underline{1.00} & {402}\,/\,{483}\,/\,{0.91} & {305}\,/\,{477}\,/\,{0.90} & {373}\,/\,\underline{493}\,/\,\underline{0.97} & 254\,/\,\underline{468}\,/\,{0.88}\\
        \midrule
        Qwen2.5-VLA-Adapter-0.5B~\cite{wang2025vlaadapter} & 10 & {353\,/\,493\,/\,0.97} & {402\,/\,475\,/\,0.90} & {214\,/\,431\,/\,0.73} & {190\,/\,473\,/\,0.89} & {195\,/\,460\,/\,0.84} \\
        \rowcolor{mypurple!20}
        \textbf{Ours} & 10 & \textbf{\underline{369}\,/\,{494}\,/\,{0.99}} & \textbf{\underline{410}\,/\,\underline{486}\,/\,\underline{0.93}} & \textbf{\underline{311}\,/\,\underline{478}\,/\,\underline{0.91}} & \textbf{\underline{379}\,/\,{491}\,/\,{0.96}} & \textbf{262\,/\,{466}\,/\,\underline{0.89}}\\
        \bottomrule
    \end{tabular}
    }
    \label{tab:unrealzoo}
\end{table*}

\section{Experiments}
\label{sec:experiments}

We evaluate our approach on EVT-Bench~\citep{wang2025trackvla}, Gym-UnrealCV~\citep{qiu2017unrealcv}, and UnrealZoo~\citep{zhong2025unrealzoo} to examine tracking performance across different environments, distractor configurations, and target appearances.

\noindent\textbf{Benchmarks.}
\textit{EVT-Bench}~\cite{wang2025trackvla} is a Habitat-based benchmark spanning 804 scenes and 100 humanoid avatars, with 25,986 episodes in total.
We use its three test splits: Single-Target Tracking (STT), Distracted Tracking (DT), and Ambiguity Tracking (AT), each containing 1,405 episodes.
These settings evaluate basic following, language-guided target discrimination, and tracking among visually ambiguous targets.

\textit{Gym-UnrealCV}~\cite{qiu2017unrealcv} measures zero-shot transfer through three scenarios: Single Target, Distractor, and Unseen Objects.
Episodes are limited to 500 control steps, and tracking is evaluated within the standard $90^\circ$ sector with a range of $7.5\,\mathrm{m}$.
This benchmark tests whether the learned policy can sustain tracking under unfamiliar environments, distractors, and target appearances.

\textit{UnrealZoo}~\cite{zhong2025unrealzoo} extends the evaluation to photorealistic virtual environments with diverse scene layouts and occlusions.
We assess single-target following using the Gym-UnrealCV tracking protocol, focusing on sustained tracking across different scenes.

\noindent\textbf{Implementation Details.}
Our model is built upon Qwen2.5-VLA-Adapter-0.5B~\cite{wang2025vlaadapter}, which combines the Prismatic VLM framework with Qwen2.5-0.5B and employs the bridge-attention mechanism from VLA-Adapter.
Target positions are encoded by their bearing angles and radial distances in the robot coordinate frame.
The policy operates in a three-dimensional continuous action space comprising forward, lateral, and angular velocities.
Training takes approximately five hours on two NVIDIA RTX 5090 GPUs.
Additional implementation details are provided in the Supplementary Material.

\subsection{Comparison with State-of-the-Art Models}
\label{sec:sota_comparison}

\noindent\textbf{Performance on EVT-Bench.}
As shown in Table~\ref{tab:evt-bench}, our method consistently achieves higher SR and TR across all three subtasks than traditional methods without VLMs.
SR reaches $87.9\%$, $41.2\%$, and $55.4\%$ on STT, DT, and AT, respectively.
With a $0.5$B backbone, it achieves competitive or better SR and TR on STT and AT compared with methods based on 7B-scale multimodal models.
However, it remains behind several of these methods on DT.
This gap may partly reflect differences in language understanding and language-related training data, which are particularly relevant to the fine-grained target descriptions in DT.

Compared with BWM~\cite{yao2026bipartite}, which uses a similarly sized backbone, our method improves SR and TR across all three subtasks.
Geometry--Progress jointly models path geometry and single-step travel distance, retaining uncertainty in both route selection and motion progress.
We then use action-induced robot pose changes to express the predicted motion in the next robot frame, providing geometrically grounded conditions for latent state transitions.
This separates target displacement from changes in the observation frame before latent prediction, reducing the need to infer their geometric relationship solely from visual features.
Future visual feature reconstruction further supervises the motion predictor through this transition branch, linking motion annotations with subsequent observations.
Together, these mechanisms provide predictive motion cues for tracking decisions.
Nevertheless, CR remains higher than that of BWM~\citep{yao2026bipartite}, particularly on DT and AT, indicating room for improvement in collision avoidance.

\noindent\textbf{Zero-Shot Performance on Gym-UnrealCV.}
Table~\ref{tab:gym-unrealcv} reports results without additional training on this benchmark.
Our model achieves EL of $500$ and SR of $1.00$ in both Single Target and Unseen Objects, matching the best reported results.
It sustains tracking over the full $500$-step horizon in all $100$ evaluation episodes per setting, despite unfamiliar environments and target appearances.
In Distractor, it reaches SR of $0.84$ and EL of $466$, compared with $0.64$ and $397$ for the backbone and $0.83$ and $461$ for BWM~\cite{yao2026bipartite}.
Despite these improvements, it remains below the 7B-scale TrackVLA~\cite{wang2025trackvla} and TrackVLA++~\cite{liu2025trackvla++} in this setting.

On Unseen Objects, SR increases from $0.35$ for OpenTrackVLA~\citep{opentrackvla2025} to $1.00$, with EL increasing from $232$ to $500$.
Geometry--Progress represents motion using geometric quantities whose meanings remain consistent across target appearances, while ego-motion conditioning accounts for viewpoint changes through explicit coordinate transformations.
Future feature reconstruction links these conditions to observed dynamics, encouraging the reuse of motion cues across visual domains.
These properties help explain the observed zero-shot generalization to unfamiliar targets and environments.

\noindent\textbf{Performance on UnrealZoo.}
Table~\ref{tab:unrealzoo} evaluates tracking across ten environments with diverse layouts and occlusions, with four random instruction changes per episode.
Compared with the backbone, our model improves AR in all ten environments and SR in eight, matching it in the remaining two.
Mean AR increases from $247.3$ to $336.0$, EL from $460.9$ to $481.9$, and SR from $0.833$ to $0.930$, a gain of $9.7$ percentage points.
The lowest scene-level SR also rises from $0.63$ to $0.89$, showing improvements in environments where the backbone performs poorly.
Compared with BWM~\cite{yao2026bipartite}, our model obtains higher AR in eight environments, with mean AR increasing from $333.1$ to $336.0$ and mean SR from $0.926$ to $0.930$.
The path--progress representation provides a common geometric description across scene layouts, and its joint distribution preserves alternative routes and progress when several motion continuations are possible.
Ego-motion conditioning accounts for viewpoint changes through coordinate relations shared across environments, while future feature reconstruction trains these motion cues to support prediction of subsequent observations.
The gains across environments support improved multi-scene tracking generalization and robustness, although the average advantage over BWM~\citep{yao2026bipartite} remains modest.

\subsection{Further Remarks}

\begin{table}[t]
    \centering
    \definecolor{ablationAccent}{HTML}{5A4487}
    \definecolor{ablationTint}{HTML}{F0ECF8}
    \definecolor{ablationMuted}{HTML}{92929C}

    \caption{
        \textbf{Component ablation on EVT-Bench STT.}
        PMP denotes probabilistic motion prediction, and WM denotes
        world-model learning. All metrics are percentages.
        \textbf{Bold} indicates the best results.
    }
    \label{tab:ablation}


    \resizebox{0.8\columnwidth}{!}{%
    \begin{tabular}{lccrrr}
        \toprule
        \multirow{2}{*}{\textbf{Variant}}
        & \multicolumn{2}{c}{\textbf{Components}}
        & \multicolumn{3}{c}{\textit{Single-Target Tracking}} \\
        \cmidrule(lr){2-3} \cmidrule(lr){4-6}
        & PMP & WM & SR$\uparrow$ & TR$\uparrow$ & CR$\downarrow$ \\
        \midrule
        Backbone
        & \textcolor{ablationMuted}{\ding{55}}
        & \textcolor{ablationMuted}{\ding{55}}
        & 52.2 & 59.1 & 30.10 \\
        (a) w/o WM
        & \textcolor{ablationAccent}{\ding{51}}
        & \textcolor{ablationMuted}{\ding{55}}
        & 80.4 & 81.8 & 9.06 \\
        (b) w/o PMP
        & \textcolor{ablationMuted}{\ding{55}}
        & \textcolor{ablationAccent}{\ding{51}}
        & 81.6 & 83.3 & 9.28 \\
        \midrule
        \rowcolor{ablationTint}
        \textbf{Ours}
        & \textcolor{ablationAccent}{\ding{51}}
        & \textcolor{ablationAccent}{\ding{51}}
        & \textbf{87.9} & \textbf{86.6} & \textbf{4.57} \\
        \bottomrule
    \end{tabular}%
    }
\end{table}

\begin{table}[t]
    \centering
    \definecolor{ablationAccent}{HTML}{5A4487}
    \definecolor{ablationTint}{HTML}{F0ECF8}
    \definecolor{ablationMuted}{HTML}{92929C}

    \caption{
        \textbf{Ablation of Geometry--Progress modeling on STT.}
        Pos.: direct position prediction;
        GP: Geometry--Progress;
        Prob.: probabilistic prediction.
        The WM baseline corresponds to the world-model-only variant
        in Table~\ref{tab:ablation}.
        \textbf{Bold} indicates the best results.
    }
    \label{tab:ablation_gp}


    \resizebox{0.9\columnwidth}{!}{%
    \begin{tabular}{lcccrrr}
        \toprule
        \multirow{2}{*}{\textbf{Variant}}
        & \multicolumn{3}{c}{\textbf{Motion prediction}}
        & \multicolumn{3}{c}{\textit{Single-Target Tracking}} \\
        \cmidrule(lr){2-4} \cmidrule(lr){5-7}
        & Pos. & GP & Prob.
        & SR$\uparrow$ & TR$\uparrow$ & CR$\downarrow$ \\
        \midrule

        WM baseline
        & \textcolor{ablationMuted}{\ding{55}}
        & \textcolor{ablationMuted}{\ding{55}}
        & \textcolor{ablationMuted}{\ding{55}}
        & 81.6 & 83.3 & 9.28 \\

        (a) Position only
        & \textcolor{ablationAccent}{\ding{51}}
        & \textcolor{ablationMuted}{\ding{55}}
        & \textcolor{ablationMuted}{\ding{55}}
        & 81.8 & 83.9 & 5.87 \\

        (b) Det. GP
        & \textcolor{ablationMuted}{\ding{55}}
        & \textcolor{ablationAccent}{\ding{51}}
        & \textcolor{ablationMuted}{\ding{55}}
        & 83.7 & 84.8 & 5.87 \\

        \midrule
        \rowcolor{ablationTint}
        \textbf{Ours}
        & \textcolor{ablationMuted}{\ding{55}}
        & \textcolor{ablationAccent}{\ding{51}}
        & \textcolor{ablationAccent}{\ding{51}}
        & \textbf{87.9} & \textbf{86.6} & \textbf{4.57} \\

        \bottomrule
    \end{tabular}%
    }
\end{table}

\noindent\textbf{Ablation Study of Motion Prediction and World Modeling.}
Table~\ref{tab:ablation} examines the contributions of probabilistic motion prediction (PMP) and world-model learning (WM) on EVT-Bench STT.
Adding PMP to the backbone increases SR from $52.2\%$ to $80.4\%$, while WM alone achieves $81.6\%$.
Combining both yields the best SR of $87.9\%$, TR of $86.6\%$, and CR of $4.57\%$, with SR gains of $7.5$ and $6.3$ percentage points over the PMP-only and WM-only variants, respectively.
These results demonstrate the complementary contributions of PMP and WM to tracking performance.
PMP supplies the policy with structured future-motion cues, while WM constrains the shared state through future visual feature reconstruction.
Conditioning the state transition on predicted motion connects these learning objectives, enabling motion prediction and future-state learning to jointly support tracking control.

\noindent\textbf{Geometry--Progress and Probabilistic Prediction.}
Starting from the WM-only baseline, Table~\ref{tab:ablation_gp} compares different target-motion representations.
Direct position prediction gives only marginal SR and TR gains, although it reduces CR from $9.28\%$ to $5.87\%$.
Compared with direct position prediction, deterministic Geometry--Progress improves SR and TR by $1.9$ and $0.9$ percentage points, respectively, with unchanged CR.
Probabilistic modeling further improves SR and TR by $4.2$ and $1.8$ percentage points and reduces CR by $1.30$ percentage points.
The deterministic comparison supports describing future motion through complementary route geometry and single-step progress, which provide richer motion information than a single future position.
The additional gains from probabilistic modeling support retaining uncertainty in routes and progress to provide the policy with multiple  future-motion hypotheses.

\begin{figure}[t]
    \centering
    \includegraphics[width=\linewidth]{
        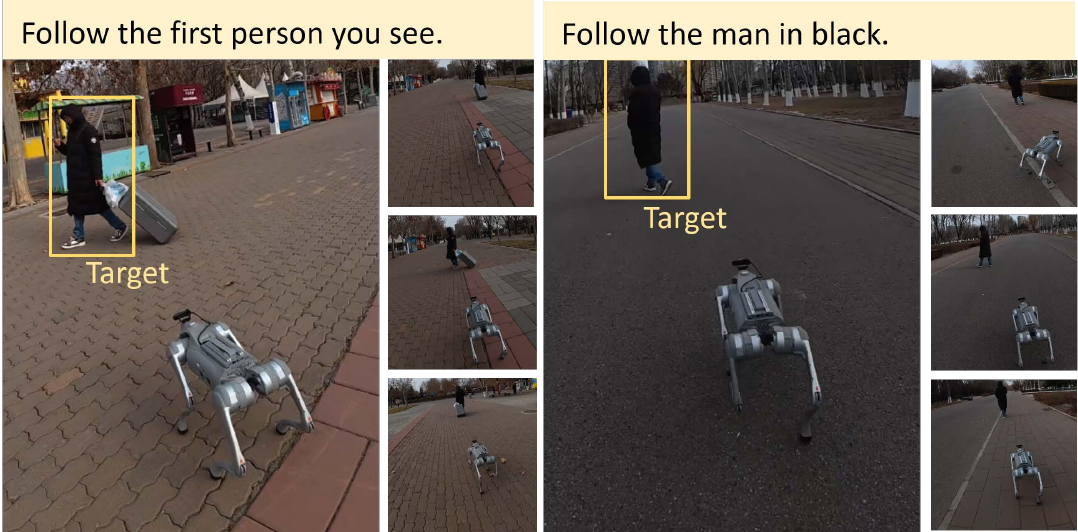
    }
    \caption{\textbf{Experiments on Real Robots.} Real-world deployment on the Unitree GO2 quadruped robot. Given natural language instructions, our method enables stable target-following in outdoor environments.}
    \label{fig:real_world}
\end{figure}

\subsection{Real-World Experiments}
\label{sec:real_world_experiments}

We deploy our motion-conditioned world model on a Unitree GO2 quadruped robot to assess its applicability to real-world target tracking.
Given sequential visual observations and a natural-language target description, the policy generates tracking actions from the current latent state and predicted target-motion representation.
The robot is equipped with an Intel RealSense D435i RGB-D camera, whose RGB observations serve as visual inputs.
The policy is trained entirely in simulation and deployed without additional real-world training or fine-tuning, enabling zero-shot sim-to-real evaluation.
Figure~\ref{fig:real_world} presents qualitative tracking sequences.
The robot continuously follows the designated target in outdoor environments.
Geometry--Progress provides predictive motion cues that complement visual appearance, while ego-motion-conditioned world-state learning accounts for viewpoint changes induced by robot movement.
Together, these components are designed to support tracking under diverse motion patterns and viewing conditions during real-world deployment.    
\section{Conclusion}
\label{sec:conclusion}
We present Motion-Conditioned World Modeling for embodied visual tracking, which incorporates predicted target-motion distributions and robot ego-motion into latent state transitions.
Geometry--Progress jointly models uncertainty in geometric paths and single-step travel distance, while coordinate transformations express predicted paths in the next robot frame to account for observer motion.
Future visual feature reconstruction supervises both state transitions and motion prediction, linking explicit motion conditions to subsequent observations.
Experiments on EVT-Bench, Gym-UnrealCV, and UnrealZoo demonstrate improved tracking and cross-environment generalization with a $0.5$B backbone, complemented by zero-shot deployment on a Unitree GO2 robot.
Future work will focus on longer-horizon motion-conditioned prediction, collision-aware control, and enhanced spatial perception of relationships among the robot, target, and surrounding environment to improve tracking robustness and safety in complex scenes with clutter and occlusion.

{
    \small
    \bibliographystyle{ieeenat_fullname}
    \bibliography{main}
}
\end{document}